\documentclass[letterpaper]{article} 
\usepackage{aaai25}  
\usepackage{times}  
\usepackage{helvet}  
\usepackage{courier}  
\usepackage[hyphens]{url}  
\usepackage{graphicx} 
\usepackage{natbib}  
\usepackage{caption} 
\usepackage{algorithm}
\usepackage{algorithmic}
\usepackage{booktabs} 
\usepackage{multirow}
\usepackage{xcolor}
\usepackage{colortbl}
\colorlet{colorSep}{blue!5}
\usepackage{amsmath} 
\usepackage{newfloat}
\usepackage{listings}
\DeclareCaptionStyle{ruled}{labelfont=normalfont,labelsep=colon,strut=off} 
\floatstyle{ruled}
\newfloat{listing}{tb}{lst}{}
\floatname{listing}{Listing}
\title{Sensing Surface Patches in Volume Rendering for Inferring Signed Distance Functions}
\author {
    Sijia Jiang\textsuperscript{\rm },
    Tong Wu\textsuperscript{\rm },
    Jing Hua\textsuperscript{\rm },
    Zhizhong Han\textsuperscript{\rm }
}
\affiliations {
    \textsuperscript{\rm }Department of Computer Science, Wayne State University, Detroit, MI, USA\\
    sijiajiang@wayne.edu, tongwu@wayne.edu, 
    jinghua@wayne.edu, h312h@wayne.edu

}

\begin{document}

\maketitle

\begin{abstract}
It is vital to recover 3D geometry from multi-view RGB images in many 3D computer vision tasks. The latest methods infer the geometry represented as a signed distance field by minimizing the rendering error on the field through volume rendering.
However, it is still challenging to explicitly impose constraints on surfaces for inferring more geometry details due to the limited ability of sensing surfaces in volume rendering. To resolve this problem, we introduce a method to infer signed distance functions (SDFs) with a better sense of surfaces through volume rendering. Using the gradients and signed distances, we establish a small surface patch centered at the estimated intersection along a ray
by pulling points randomly sampled nearby. Hence, we are able to explicitly impose surface constraints on the sensed surface patch, such as multi-view photo consistency and supervision from depth or normal priors, through volume rendering. We evaluate our method by numerical and visual comparisons on scene benchmarks. Our superiority over the latest methods justifies our effectiveness. Our code is available at \url{https://github.com/MachinePerceptionLab/Surface-Sensing-SDF}.
\end{abstract}

\section{Introduction}
3D reconstruction from multi-view images is an important task in 3D computer vision. Classic methods like structure from motion and multi-view stereo~\cite{schoenberger2016sfm,schoenberger2016mvs} estimate 3D point clouds per the multi-view photo consistency. With deep learning models~\cite{yao2018mvsnet,Yu2022MonoSDF}, we are able to learn depth priors from a large scale dataset, which can be further generalized to infer depth maps from unseen images. Although these methods can estimate a coarse geometry from depth predictions, it is still a challenge to recover continuous and complete surfaces with details from multi-view images.

The latest methods~\cite{GEOnEUS2022,Yu2022MonoSDF,wang2022neuris,guo2022manhattan} infer signed distance functions (SDFs) from multi-view images through volume rendering, and then run the marching cubes~\cite{Lorensen87marchingcubes} to reconstruct the surface. To supervise signed distances, they transform the predicted signed distances into radiance to render a pixel color by integrating colors along a ray, which can be optimized by minimizing the difference to the ground truth pixel color. Although these methods obtained smooth and complete reconstructions, severe artifacts may appear in the empty space, the reconstructed surfaces may drift away from the GT surface, and few details can be revealed on the surface, due to the unawareness of the surface. Hence, how to sense the surface and further impose more effective surface constraints along with the volume rendering is the key to improve the learning of SDF.

To resolve this problem, we propose to infer an SDF from multi-view images through volume rendering with a better sense of surface. Our novelty lies in the way of establishing a surface patch around an estimated ray-surface intersection, which enables to explicitly impose more effective constraints on surface patches, along with the volume rendering.
Specifically, using the predicted signed distances and gradients, we randomly sample queries near an estimated intersection, and pull them onto the zero level set, which produces a surface patch. With the sensed surface patch, we are able to explicitly impose surface constraints such as multi-view photo consistency and supervision from depth or normal priors, to improve the SDF inference through volume rendering. We justify the effectiveness of our modules, and report superiority performance over the latest methods in terms of numerical and visual comparisons on widely used benchmarks. Our contributions are listed below.

\begin{itemize}
\item We introduce a method for SDF inference that can get constrained not only through volume rendering but also by constraints that can be explicitly imposed on surfaces. It significantly improves the accuracy of inferred SDFs.

\item We propose to use predicted signed distances and gradients to sense a surface patch near the estimated intersection of a ray and the zero level set in volume rendering.
\item We justify the feasibility of our approach and report the state-of-the-art performance on the widely used benchmarks.
\end{itemize}

\section{Related Work}
\noindent\textbf{Mult-view 3D Reconstruction. }3D shape reconstruction from multiple images has been extensively studied~\cite{schoenberger2016sfm,schoenberger2016mvs,mildenhall2020nerf,Vicini2022sdf}. Given multiple RGB images, classic multi-view stereo (MVS)~\cite{schoenberger2016sfm,schoenberger2016mvs} employ multi-view photo consistency to estimate depth maps. 
However, these methods are limited by large viewpoint variations and complex illumination. Alternatively, with multiple silhouette images,~\cite{273735visualhull} proposed to reconstruct 3D shapes as voxel grids using space carving. These methods lack the ability to reveal concave structures and work with high-resolution voxel grids.

Recent methods~\cite{yao2018mvsnet} employ neural networks to learn prior knowledge to predict depth maps. During training, they learn priors using depth supervision or multi-view consistency in an unsupervised way~\cite{zhizhongiccv2021completing,Jiang2019SDFDiffDR,handrwr2020}, and then generalize the learned priors to predict depth images for unseen cases through a forward pass. Recent works ~\cite{Huang_2024,chen2023neusgneuralimplicitsurface,yu2024gaussianopacityfieldsefficient,guédon2023sugarsurfacealignedgaussiansplatting,zhang2024gspull} leverage 3D Gaussian Splatting for surface reconstruction. However, they are inferior to neural implicit methods in quality due to the explicit and disconnected 3D Gaussians.

These methods reconstructed 3D shapes as point clouds or voxel grids, both of which are discrete 3D representations. While neural implicit representations represent continuous surfaces as the zero level set for 3D reconstruction.

\noindent\textbf{Neural Implicit Representations. }
Neural implicit representations have become a popular 3D representation, using coordinate-based neural networks to map coordinates to signed distances or occupancy labels. These can be inferred from 3D supervision~\cite{takikawa2021nglod,Liu2021MLS,tang2021sign}, point clouds~\cite{Zhou2022CAP-UDF,chaompi2022,ChaoSparse,Baoruicvpr2023,chao2023gridpull,localn2nm2024,ma2023learning,zhou2023levelset,multigrid}, or multi-view images~\cite{mildenhall2020nerf,guo2022manhattan,zhang2024learning,Hu2023LNI-ADFP,sijia2023quantized}. Methods using 3D supervision or point clouds typically skip positional encodings, while multi-view approaches use them to capture high-frequency details.

Differentiable rendering enables tuning implicit representations by minimizing errors between the rendered and ground truth images. Surface rendering methods like DVR~\cite{DVRcvpr} and IDR~\cite{yariv2020multiview} predict radiance on surfaces and use view direction for high-frequency detail but require background filtering. Volume rendering methods like NeRF~\cite{mildenhall2020nerf} and its variations~\cite{mueller2022instant,Azinovic_2022_CVPR} model geometry and color without masks. UNISURF~\cite{Oechsle2021ICCV} and NeuS~\cite{neuslingjie} refine occupancy and signed distance fields using revised rendering equations, with improvements via depth~\cite{Yu2022MonoSDF}, normals~\cite{guo2022manhattan}, and multi-view consistency~\cite{GEOnEUS2022}.

Our approach differs previous methods by sensing surface patches along rays and imposing explicit surface constraints to recover finer geometric details.

\section{Method}
\noindent\textbf{Overview. }We aim to recover the geometry of a scene by learning an SDF $f$ from $K$ posed images $C_k^{GT}$. We can use additional supervision such as depth $D_k^{GT}$ and normal $N_k^{GT}$ maps that are either captured by real sensors or estimated by monocular networks, where $k\in[1,K]$. For randomly sampled 3D queries $q$, $f$ predicts its signed distance $f(q)$ at $q$. We parameterize the SDF $f$ using a coordinate-based MLP with parameters $\theta$.

\begin{figure*}[t]
  \centering
   \includegraphics[width=0.85\linewidth]{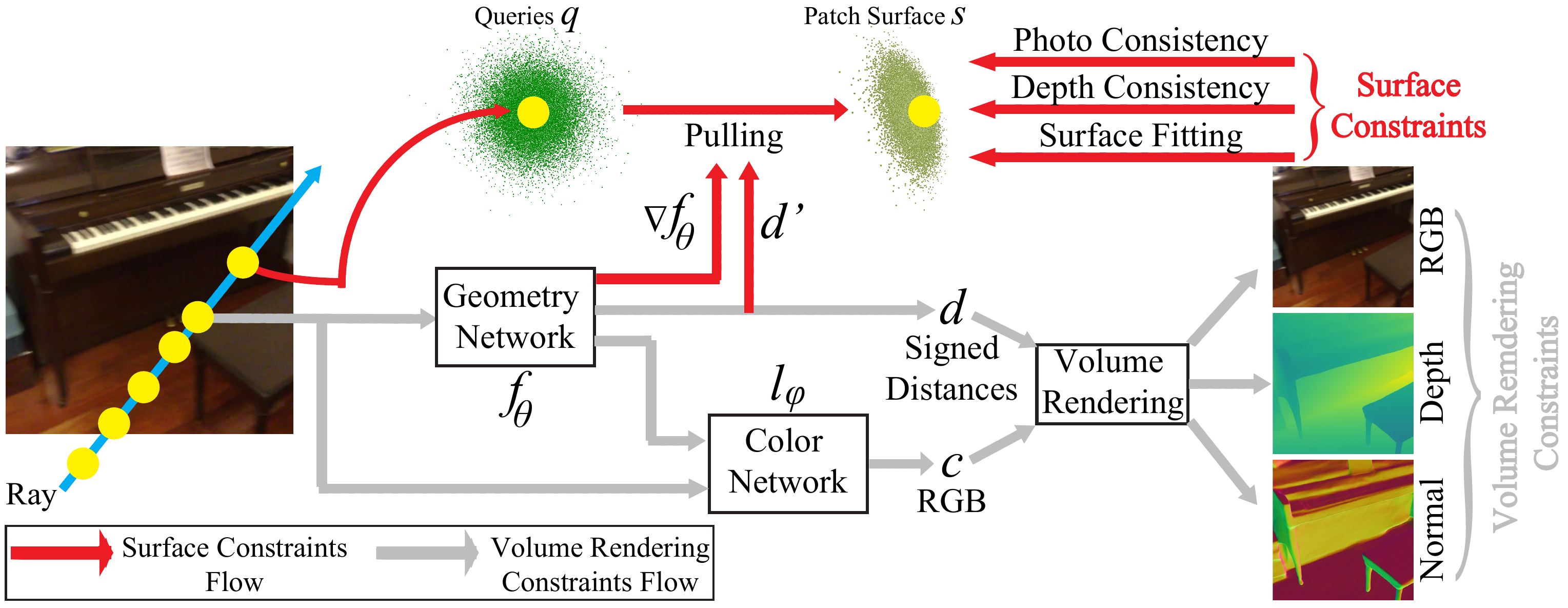}
\caption{\label{fig:overview} Overview of our method. We infer SDF $f_{\theta}$ from multi-view images including RGB images, depth and normal maps that were either captured by sensors or estimated by monocular networks. Using the predicted signed distances and gradients $\nabla f_{\theta}$, we are enabled to sense a surface patch $s$ by pulling randomly sampled queries $q$ onto the zero level set as shown in Fig.~\ref{fig:surface} (d). With $s$, we can infer $f_{\theta}$ using both supervision through volume rendering and constraints that can be explicitly imposed on the sensed surface $s$.}                  
\end{figure*}

As illustrated in Fig.~\ref{fig:overview}, we aim to infer $f_{\theta}$ along with a color function $l_{\phi}$ which is parameterized by another MLP with parameters $\phi$ through volume rendering. Our optimization is to minimize a loss $L$ according to $\{C_k^{GT},D_k^{GT},N_k^{GT}\}$,

\begin{equation}
\label{eq:objective}
\min_{\theta,\phi}L(f_{\theta},l_{\phi},\{C_k^{GT}\},\{D_k^{GT}\},\{N_k^{GT}\}).
\end{equation}

\noindent\textbf{Geometry prediction. }We use an MLP to approximate the geometry function $f_{\theta}$. For a query $q$, $f_{\theta}$ uses the coordinate and the position encoding~\cite{mildenhall2020nerf} $e(q)$ to capture the geometry with high frequency. The SDF $f_{\theta}$ predicts the signed distance $d=f_{\theta}(q,e(q))$ at $q$.

\noindent\textbf{Color Prediction. }We use another MLP to approximate the color function $l_{\phi}$. To model the view-dependent color $c$ at $q$, we also leverage the view direction $v$, the gradient $g$ in the signed distance field, and the feature $z$ of geometry around $q$. Hence, we predict color by $c=l_{\phi}(q,e(q),v,g,z)$. We obtain the gradient $g$ from the geometry function $f_{\theta}$ as $g=\nabla f_{\theta}$, which can be produced by automatic differentiation from the geometry network. And we use the output of one FC layer from the geometry network as the feature $z$.

\noindent\textbf{Volume Rendering. }We use volume rendering to render the radiance field represented by the SDF $f_{\theta}$ and the color function $l_{\phi}$ into images. The parameters $\theta$ and $\phi$ can be learned by minimizing the rendering error to the ground truth images.

We start from emitting a ray $r$ from the camera center $o$ through a randomly sampled pixel on an image. To render a color along the ray $r$ pointing to a view direction $v$, points $\{q_i|i\in[1,I]\}$ are sampled along $r$ by $q_i=o+t_i*v$. The geometry network $d_i=f_{\theta}(q_i,e(q_i))$ and color network $c_i=l_{\phi}(q_i,e(q_i),v,g_i,z_i)$ are used to predict the signed distance and color at each sampled point $q_i$. We follow VolSDF~\cite{yariv2021volume} to transform the signed distance $d_i$ to density values $\sigma_i$ for volume rendering.

Following NeRF~\cite{mildenhall2020nerf}, the color $C_r$ for the ray $r$ is integrated by,

\begin{equation}
\label{eq:rendering}
\alpha_i=1-exp(-\delta_i\sigma_i),\,
T_i=\prod_{j=1}^{i-1}(1-\alpha_j),\,
C_r=\sum_{i=1}^IT_i\alpha_ic_i,
\end{equation}

\noindent where $\alpha_i$ is the alpha value at point $q_i$, $\delta_i$ is the interval between neighboring points, and $T_i$ is the transmittance through $q_i$. Similarly, we can render depth $D_r$ and normal map $N_r$ using the following equations,

\begin{equation}
\label{eq:depthrendering}
D_r=\sum_{i=1}^M T_i\alpha_it_i, \, N_r=\sum_{i=1}^M T_i\alpha_in_i.
\end{equation}

\begin{figure}
  \includegraphics[width=\linewidth]{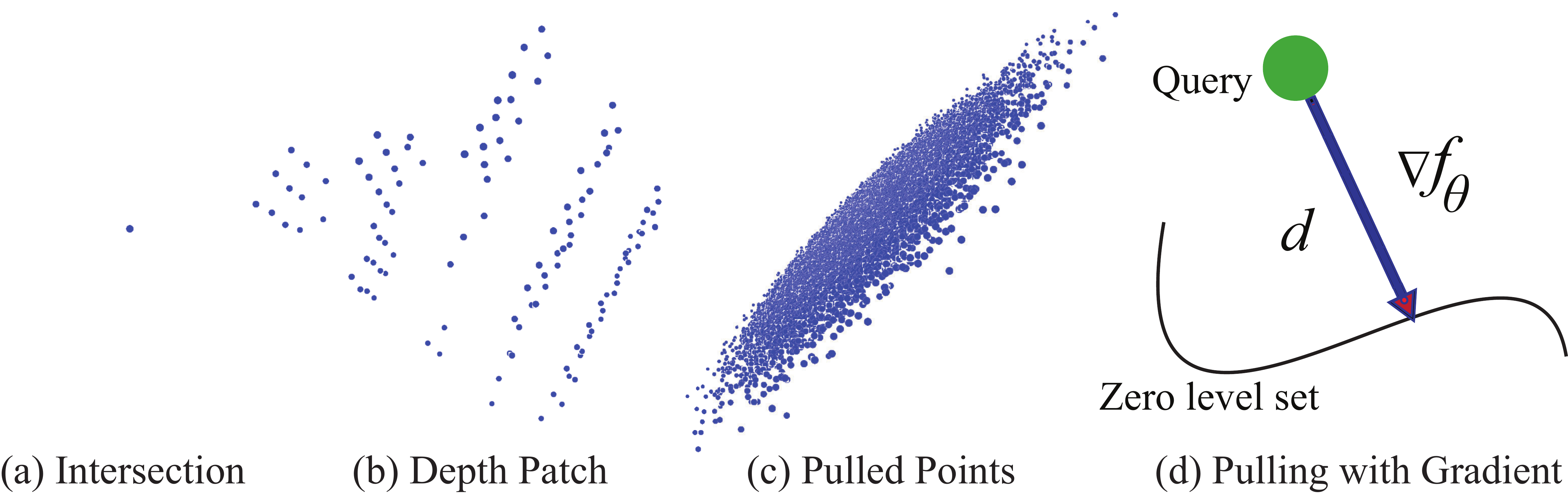}
  \caption{\label{fig:surface}Patch Difference. Current methods mainly impose constraints on single points on zero level set in (a), rather than a patch, since it is very hard to obtain a 3D surface patch during volume rendering. Different from obtaining a patch from depth in (b), our method projects randomly sampled points on the zero level set to obtain the surface patch which is more continuous and representative in (c).}
\vspace{-0.1in}
\end{figure}

\noindent\textbf{Surface Sense. }We use the geometry function $f_{\theta}$ to sense the surface. NeRF or its variations~\cite{yariv2021volume,yiqunhfSDF,wang2022neuris} apply the secant method~\cite{MeschederNetworks} to estimate the intersection of a ray and the surface per the signed distances or occupancy labels predicted at points sampled along the ray, as shown in Fig.~\ref{fig:surface} (a). Although it is a way of sensing a surface, on the zero level set, only one single point gets supervised during the differentiable rendering procedure.

We introduce a novel way of sensing a surface patch by pulling randomly sampled queries. The sensed surface is represented by pulled queries as shown in Fig.~\ref{fig:surface} (c), which is a denser and more continuous surface representation than a single point intersection in Fig.~\ref{fig:surface} (a). It is much more geometry-aware than the single point intersection and also points back-projected from a depth patch in Fig.~\ref{fig:surface} (b).

Using the character of SDF, we can leverage the gradient and the signed distance to move any points onto the zero level set~\cite{Zhizhong2021icml,chou2022gensdf,onsurfacepriors2022,predictivecontextpriors2022}.

More specifically, for a 3D point $q=[x,y,z]$, it is located on the $d$-level set in a signed distance field represented by $f_{\theta}$, where $d=f_{\theta}(q,e(q))$. The gradient at $q$ in the field $\nabla f_{\theta}(q,e(q))=[\partial f_{\theta}/\partial x,\partial f_{\theta}/\partial y,\partial f_{\theta}/\partial z]$ points to the level sets with larger signed distances than $d$. As demonstrated in Fig.~\ref{fig:surface} (d), we can pull $q$ onto its zero level set by moving it along its gradient $g$ with a stride of $|d|$,

\begin{equation}
\label{eq:projection}
q'=q-d\times\nabla f_{\theta}(q,e(q))/||\nabla f_{\theta}(q,e(q))||_2,
\end{equation}

\noindent where $q'$ is the projection of point $q$ on the zero level set.

Our approach of sensing a surface patch is to use the projections of randomly sampled points to form a point surface. Specifically, we sample a set of points $\{p_j|j\in[1,J]\}$ around a 3D anchor $q$ using a Gaussian distribution. The anchor $q$ could be either the intersection estimated along a ray or a point back-projected from posed depth maps. The Gaussian distribution has a variance $\tau^2$ to cover the area centered at $q$ between neighboring rays. $\tau^2$ determines the size of the sensed patch, we report ablation study on $\tau^2$ in experiments. Using Eq.~\ref{eq:projection}, we pull each $p_j$ to its projection $p_j'$ on the zero level set, which forms a surface patch $s$. We denote $s$ that we sense as,

\begin{equation}
\label{eq:surfacesense}
s=\{p_j'|j\in[1,J]\}.
\end{equation}

\section{Optimization}
\subsection{Constraints through Volume Rendering}
The following losses are used to provide constraints through the volume rendering. We follow MonoSDF~\cite{Yu2022MonoSDF} to use losses to supervise rendered RGB, depth and normal maps. The GT depth maps and normal maps are either captured by sensors or estimated by monocular networks. 

\noindent\textbf{RGB Rendering Loss. }We use the RGB images $C_k^{GT}$ to supervise the images $C_k$ rendered from the field using Eq.~\ref{eq:RGBRec},

\begin{equation}
\label{eq:RGBRec}
L_{RGB}=\sum\nolimits_{r\in B}||C_k(r)-C_k^{GT}(r)||_1,
\end{equation}

\noindent where $C_K^{GT}(r)$ is the ground truth color at pixel $r$ and $B$ denotes a set of sampled rays in a mini-batch.

\noindent\textbf{Depth Rendering Loss. }With GT depth maps $D_k^{GT}$, we can supervise the depth maps $D_k$ rendered from Eq.~\ref{eq:depthrendering} below,

\begin{equation}
\label{eq:depRec}
L_{DR}=\sum\nolimits_{r\in B}||D_k(r)-D_k^{GT}(r)||_1,
\end{equation}

\noindent where $D_k^{GT}(r)$ is the depth cue at the ray $r$ in mini-batch $B$.

\noindent\textbf{Normal Rendering Loss. }Given GT normal maps $N_k^{GT}$, we can also supervise the normal maps $N_k$ rendered from Eq.~\ref{eq:depthrendering},

\begin{equation}
\begin{split}
\label{eq:NorRec}
L_{Nor}=\sum\nolimits_{r\in B}||N_k(r)-N_k^{GT}(r)||_1+ \\
||1-N_k(r)^TN_k^{GT}(r)||_1,
\end{split}
\end{equation}

\noindent where $N_k^{GT}(r)$ is the normal cue at the ray $r$ in mini-batch.

\noindent\textbf{Eikonal Loss. }To learn $f_{\theta}$ as an SDF, we constrain the gradients $\nabla f_{\theta}$ in the field using the Eikonal term,

\begin{equation}
\label{eq:Eik}
L_{Eik}=\sum\nolimits_{q\in B}(||\nabla f_{\theta}(q,e(q))||_2-1)^2,
\end{equation}

\noindent where $q\in B$ denotes all points sampled on rays in the mini-batch $B$.

\subsection{Surface Constraints}
With a surface patch $s$ that we sense using Eq.~\ref{eq:surfacesense}, we are able to explicitly impose surface constraints on $s$. 

\noindent\textbf{Depth Consistency. }With GT depth maps, we can regress depth of points  $p_j'$ on the surface patch $s$  by projecting them to the current view plane. The consistency of the calculated depth and the regressed depth can be computed as follow,

\begin{equation}
\label{eq:depthregression}
L_{DC}=\sum\nolimits_{p_j'\in s}w_j\times(D_k^{GT}(p_j')-Z_k(p_j'))^2,
\end{equation}

\noindent where $D_k^{GT}(p_j')$ is the depth interpolated at the projection of $p_j'$ on the depth map $D_k^{GT}$ using bilinear interpolation, and $Z_k(p_j')$ is the projected depth from $p_j'$ to the plane of $D_k^{GT}$ using the pose and the intrinsic matrix of the camera. We use a mask $w_j$ to rule out $p_j'$ whose projections are out of the view range or $|D_k^{GT}(p_j')-Z_k(p_j')|$ is larger than $15mm$ which indicates potential invisibility from the current view.

$L_{DC}$ is different from $L_{DR}$ in Eq.~\ref{eq:depRec}. $L_{DC}$ only constrains zero level set in the field, while $L_{DR}$ constrains all locations sampled along a ray $r$, due to the integration in rendering.

\noindent\textbf{Photometric Consistency. }With the surface patch $s$, the photo consistency can be constrained on $s$ across different views. We use the normalization cross correlation (NCC) of patches in one reference gray image $U_{k1}'=gray(C_{k1}^{GT})$ and another source gray image $U_{k2}'=gray(C_{k2}^{GT})$,


\begin{equation}
\label{eq:ncc}
NCC(U_{k1}'(s),U_{k2}'(s))=\frac{Cov(U_{k1}'(s),U_{k2}'(s))}{\sqrt{Var(U_{k1}'(s))Var(U_{k2}'(s))}},
\end{equation}

\noindent where $Cov$ and $Var$ are the covariance and the variance over the gray level color interpolated at projections of $\{p_j'\}$, denoted as $U'(s)$. We regard the view to be rendered as the reference image $U_{k1}'$ and the neighboring eight images as the source images $\{U_{k2}'\}$. Considering occlusion, only the top three largest NCC scores are used to compute the following photometric consistency loss,

\begin{equation}
\label{eq:ncc1}
L_{NCC}=\frac{\sum\nolimits_{k2=1}^31-NCC(U_{k1}'(s),U_{k2}'(s))}{3}.
\end{equation}

\noindent\textbf{Surface Fitting. }We improve the smoothness of the surface patch $s$ using a surface fitting loss, and hope all $p_j'$ on $s$ can locate on the same plane determined by the depth supervision $D_k^{GT}(r)$ and the normal supervision $N_k^{GT}(r)$ of ray $r$. We use $\alpha x+\beta y+\gamma z+\mu=0$ to represent the plane, and solve $[\alpha,\beta,\gamma,\mu]$ using $D_k^{GT}(r)$ and $N_k^{GT}(r)$. We measure the fitting error using the equation below,

\begin{equation}
\label{eq:fitting}
L_{Fit}=\sum\nolimits_{p_j'\in s}w_j\times\eta_j\times||\alpha x_j+\beta y_j+\gamma z_j+\mu||^2,
\end{equation}

\noindent where $w_j$ is the mask introduced in Eq.~\ref{eq:depthregression} and $\eta_j$ is the confidence determined by the gradient $\nabla f_{\theta}(p_j',e(p_j'))$. We model the confidence as the consistency between the gradient $\nabla f_{\theta}(p_j',e(p_j'))$ and the normal $N_k^{GT}(r)$ of $r$ using the cosine distance $\eta_j=cos(\nabla f_{\theta}(p_j',e(p_j')),N_k^{GT}(r))$, which pushes $p_j'$ more onto the plane if its gradient is well aligned with the normal $N_k^{GT}(r)$ of $r$. Note that $\nabla f_{\theta}(p_j',e(p_j'))$ involves second order derivative due to the $p_j'$ in Eq.~\ref{eq:projection} which helps the network to find better solutions~\cite{DBLP:journals/corr/abs-2106-10811}.

\begin{figure}[tb]
  \centering
   \includegraphics[width=1.\linewidth]{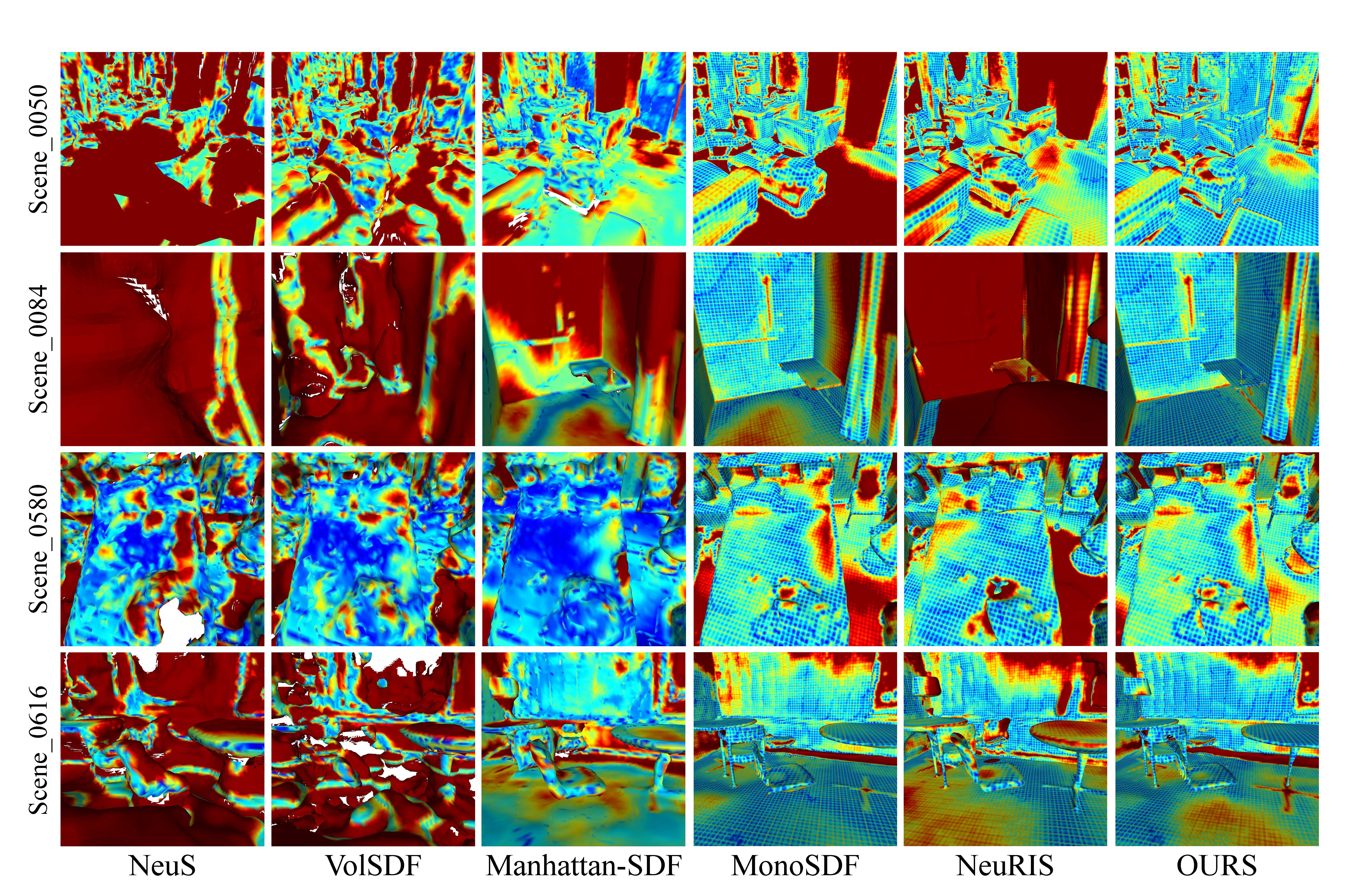}
\caption{\label{fig:scannet}Error map comparison on ScanNet (bigger error: red, smaller error: blue) highlights our superiority.}
\vspace{-0.10in}
\end{figure}

\subsection{Loss Function}
We use all these constraints to infer the geometry and color in the field below,

\begin{equation}
\label{eq:all}
\begin{split}
L=& L_{RGB}+\lambda_1L_{DR}+\lambda_2L_{Nor}+\lambda_3L_{Eik}\\
&+\lambda_4L_{DC}+\lambda_5L_{NCC}+\lambda_6L_{Fit},\\
\end{split}
\end{equation}

\noindent where $\lambda_1$ to $\lambda_6$ are balance weights which make each term contribute to the performance equally.

\begin{figure}[tb]
  \centering
   \includegraphics[width=1.0\linewidth]{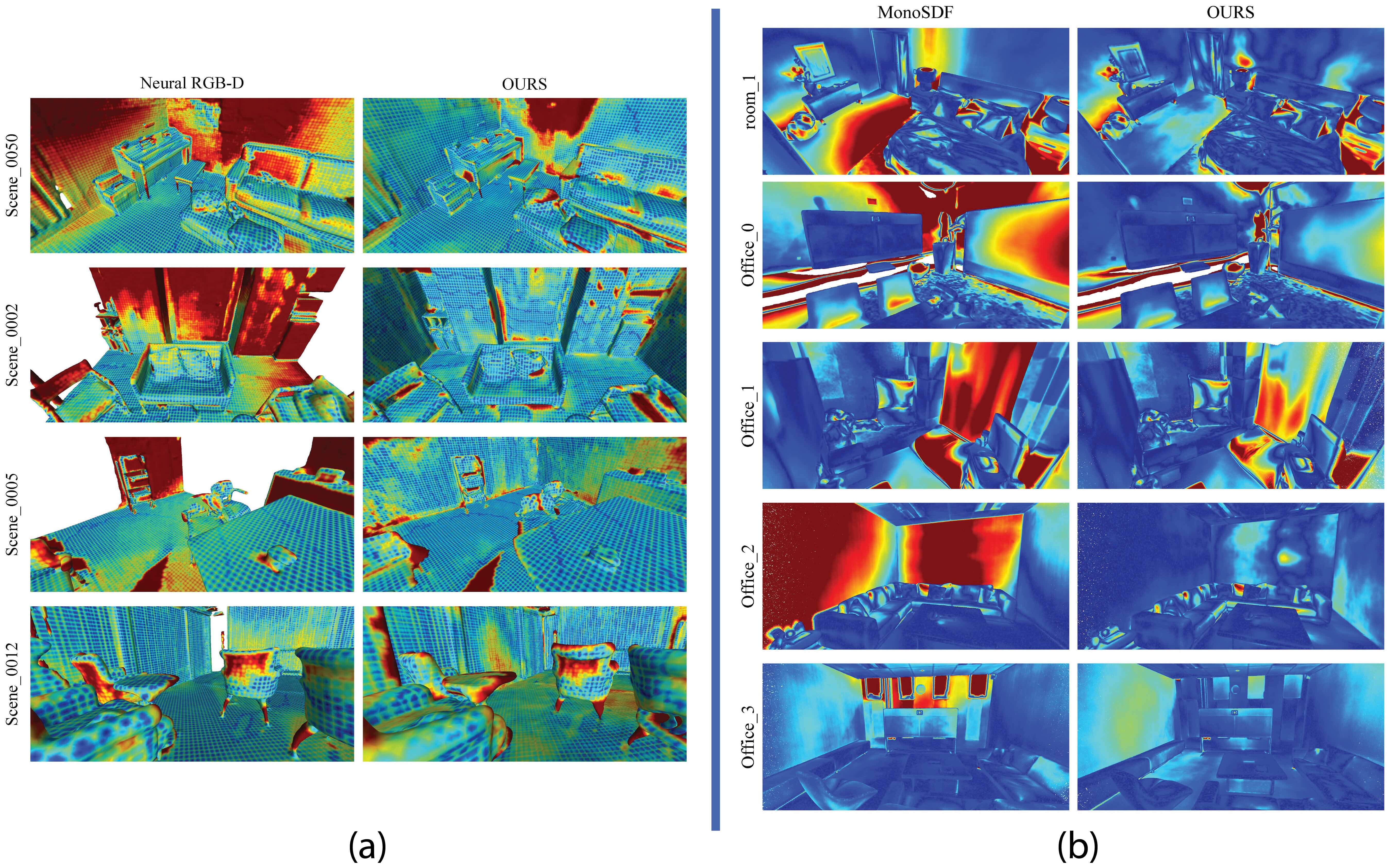}
\caption{\label{fig:scannet-replica}(a) Visual comparison on ScanNet released by NeuralRGBD. We use error maps (bigger error: red, smaller error: blue) to highlight our superiority over NeuralRGBD. (b) Visual comparison on Replica. We use error maps (bigger error: red, smaller error: blue) to highlight our superiority over MonoSDF.}
\end{figure}

\section{Experiments and Analysis}
We report numerical and visual comparisons with the latest methods on real-world indoor scenes to highlight our superiority and justify the effectiveness of modules in our method.

\begin{table}[tb]
\centering
\resizebox{\linewidth}{!}{
\begin{tabular}{c| c c c c c c}
\toprule
& Acc$\downarrow$ & Comp$\downarrow$ & CD-L1$\downarrow$ & Prec$\uparrow$ & Recall$\uparrow$ & F-score$\uparrow$\\
\midrule
COLMAP &0.047 &0.235& 0.141& 0.711& 0.441& 0.537\\
UNISURF &0.554& 0.164& 0.359& 0.212& 0.362& 0.267\\
NeuS &0.179& 0.208 &0.194& 0.313& 0.275 &0.291\\
VolSDF & 0.414 &0.120 &0.267& 0.321&0.394& 0.346\\
Manhattan & 0.072 &0.068& 0.070 &0.621& 0.586& 0.602\\
NeuRIS& 0.050& 0.049& 0.050& 0.717& 0.669 &0.692\\
Neuralangelo& 0.245& 0.272& 0.258& 0.274& 0.311 &0.292\\
MonoSDF& \textbf{0.035} &  0.048 &  0.042 & 0.799 & 0.681 & 0.733\\
\hline
Ours & 0.036 & \textbf{0.039} & \textbf{0.037} & \textbf{0.820} & \textbf{0.777} &\textbf{0.797}\\
\bottomrule
\end{tabular}}
\caption{Comparisons on ScanNet released by MonoSDF.}
\label{table:scenenetmonosdf}
\vspace{-0.2in}
\end{table}

\noindent\textbf{Datasets. }We evaluate our method by comparisons with the latest methods for scene reconstruction from multi-view images under both synthetic scenes and real scans. The synthetic indoor scenes are Replica~\cite{DBLP:journals/corr/abs-1906-05797}, released by MonoSDF~\cite{Yu2022MonoSDF}, the real scans of indoor scenes are ScanNet~\cite{DBLP:journals/corr/DaiCSHFN17}, released by either MonoSDF or NeualRGBD~\cite{Azinovic_2022_CVPR}, and real-world large-scale indoor scenes, Tanks and Temples~\cite{Knapitsch2017}, released by MonoSDF. 

\noindent\textbf{Baselines. }We compare our method with SOTA neural implicit-based reconstruction methods, including COLMAP~\cite{schoenberger2016sfm}, UNISURF~\cite{Oechsle2021ICCV}, NeuS~\cite{neuslingjie}, VolSDF~\cite{yariv2021volume}, Manhattan-SDF~\cite{guo2022manhattan}, NeuRIS~\cite{wang2022neuris}, Neuralangelo~\cite{li2023neuralangelo}, and MonoSDF on ScanNet, using MonoSDF’s experimental setup with monocular depth and normal cues as supervision. For comparisons with NeuralRGBD, we adopt its setting, using sensor-captured depth for fair evaluation.


\noindent\textbf{Evaluation Metrics. }As for evaluation metrics, we follow previous methods~\cite{Yu2022MonoSDF,Azinovic_2022_CVPR}, and report accuracy, completeness, Chamfer Distance (CD), the F-score with a threshold of 5cm, Precision and Recall, as well as normal consistency (NC). 

\noindent\textbf{Details. }For each posed view, we sample 1024 rays per training batch. Using VolSDF's error-bounded sampling strategy and architecture, we sample points along rays. To create surface patches, geometry cues are backprojected (either predicted monocular or dataset-provided sensor cues) to obtain 3D anchor points \(q\). Around each anchor, we define an isotropic Gaussian distribution \(N(q, \tau^2)\) and sample \(J=9\) points, with \(\tau^2\) controlling patch size. Ablation studies in Tab.~\ref{table:pointnumber} analyze the effect of different \(\tau^2\) values. Loss weights are set as \(\lambda_1=0.1\), \(\lambda_2=0.05\), \(\lambda_3=0.05\), \(\lambda_4=0.5\), and \(\lambda_6=0.5\) for balanced contributions. To improve coarse shape reconstruction, we apply inverse weight annealing for \(L_{NCC}\), setting \(\lambda_5=0\) for the first 100 epochs and gradually increasing it to 0.1.

\begin{table}[tb]
\centering
\resizebox{\linewidth}{!}{
\begin{tabular}{c c| c c c c c}
\toprule
& &Scene\textunderscore0050 & Scene\textunderscore0002 & Scene\textunderscore0005 & Scene\textunderscore0012 & Mean\\
\midrule
{\multirow{6}{*}{\rotatebox[origin=c]{90}{NeuralRGBD}}}
&Acc[cm]$\downarrow$ &2.84 &3.93 &\textbf{4.13} &2.78 &\textbf{3.42}\\
&Comp[cm]$\downarrow$&10.41 &32.72& 33.35& 3.03&19.88 \\
&Chamfer-L1$\downarrow$ &6.63 &18.33 &18.74 &2.90 &11.65 \\
&Prec$\uparrow$ & 90.24 &73.74 &72.68 &91.33 & 82.00\\
&Recall$\uparrow$ & 71.19 &34.05 &41.30 &86.33 & 58.22\\
&F-score$\uparrow$ & 79.59 &46.59 &52.67 &88.76 & 66.90\\
\hline
{\multirow{6}{*}{\rotatebox[origin=c]{90}{OURS}}}
&Acc[cm]$\downarrow$ &\textbf{2.78} &\textbf{3.70} &5.51 &\textbf{2.71} &3.68\\
&Comp[cm]$\downarrow$ &\textbf{3.16} &\textbf{7.10} & \textbf{8.98} & \textbf{2.83}& \textbf{5.52}\\
&Chamfer-L1$\downarrow$ &\textbf{2.97} &\textbf{5.4} &\textbf{7.25} &\textbf{2.77} & \textbf{4.60}\\
&Prec$\uparrow$ & \textbf{92.54} &\textbf{86.90} &\textbf{80.66} &\textbf{91.48} &\textbf{87.89}\\
&Recall$\uparrow$ & \textbf{85.27} &\textbf{73.12} &\textbf{73.83} &\textbf{88.33} &\textbf{80.14}\\
&F-score$\uparrow$ & \textbf{88.76} &\textbf{79.41} &\textbf{77.09} &\textbf{89.88} &\textbf{83.79}\\
\bottomrule
\end{tabular}}
\caption{Numerical comparison with NeuralRGBD on ScanNet subsets used by NeuralRGBD. GT meshes were generated with TSDF fusion using the same amount of images and depth maps as did NeuralRGBD on each scene. Under the same setting as NeuralRGBD, GT depth is sensor captured depth maps.}
\label{table:scenenetRGBD}
\end{table}

\begin{figure}
\centering
\includegraphics[width=\linewidth]{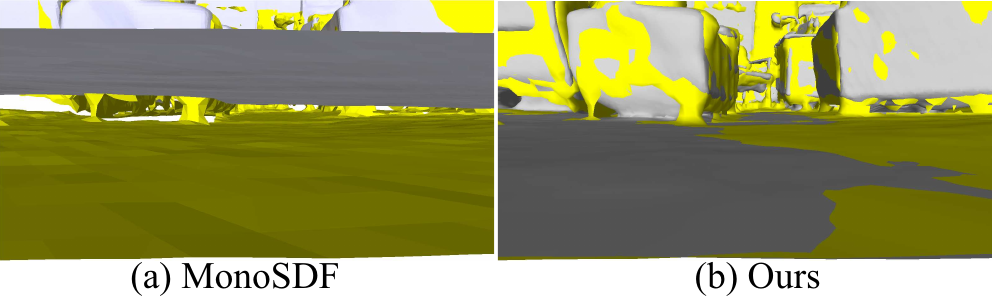}
\caption{\label{fig:ground}Compactness comparison with MonoSDF (GT mesh:Gray).}
\vspace{-0.1in}
\end{figure}

\subsection{Experimental Results}
\noindent\textbf{ScanNet from MonoSDF. }Numerical comparisons with the latest methods are presented in Tab.~\ref{table:scenenetmonosdf}. Following MonoSDF~\cite{Yu2022MonoSDF}, we use monocular cues estimated by the pretrained Omnidata model~\cite{eftekhar2021omnidata}. To align the Omnidata depth (range $[0,1]$) with the rendered depth $D_r$ in Eq.~\ref{eq:depRec}, we solve scale $w$ and shift $q$ parameters via least-squares optimization per mini-batch. These parameters are used to backproject depth and calculate $L_{DC}$ in Eq.~\ref{eq:depthregression}.  

As shown in Tab.~\ref{table:scenenetmonosdf}, our method produces more accurate and smoother surfaces. Neuralangelo struggles with real-world indoor scenes like ScanNet, even with parameter tuning and monocular cues, as evident in our visual comparisons (Fig.~\ref{fig:scannet}) and error maps (supplementary). Additionally, our method enhances the quality of volume-rendered images, with visual and PSNR comparisons available in the supplementary material.

\begin{table}[tb]
\centering
\resizebox{\linewidth}{!}{
\begin{tabular}{c c c c c c c}
\toprule
&  & Test split & & & Train split  \\
& Normal C.$\uparrow$ & CD-L1$\downarrow$ & F-score$\uparrow$ & Normal C.$\uparrow$ & CD-L1$\downarrow$ & F-score$\uparrow$\\
\midrule
MonoSDF & \textbf{92.11} &  2.94 &  86.18 & 93.86 & 2.63 & 92.12\\
\hline
Ours & 91.68 & \textbf{2.81} & \textbf{89.73} & \textbf{94.29} & \textbf{2.37} & \textbf{94.09}\\
\bottomrule
\end{tabular}}
\caption{Comparisons with MonoSDF on Replica.}
\label{table:replica}
\end{table}


\noindent\textbf{ScanNet from NeuralRGBD. }NeuralRGBD~\cite{Azinovic_2022_CVPR} evaluated on a different subset of ScanNet using sensor-captured depth maps. In this case, we use only the ground truth depth maps for $L_{DR}$ in Eq.~\ref{eq:depRec} and $L_{DC}$ in Eq.~\ref{eq:depthregression}. Tab.~\ref{table:scenenetRGBD} demonstrates that our method outperforms the latest approaches, and Fig.~\ref{fig:scannet-replica} (a) highlights our more accurate reconstruction, especially on planar structures.


\noindent\textbf{Replica. }We evaluate our method on synthetic Replica scenes, following the MonoSDF setup and using monocular cues predicted by the pretrained Omnidata model. The scale and shift are solved in Tab.~\ref{table:scenenetmonosdf}. Tab.~\ref{table:replica} shows our superiority over MonoSDF, both with and without Replica pretraining ("Train split" and "Test split"). Additionally, we compare against SOTA dense monocular SLAM methods (DROID-SLAM~\cite{teed2021droid}, NICER-SLAM~\cite{nicerslam}) and the RGB-D SLAM system NICE-SLAM~\cite{Zhu2021NICESLAM}, retrained under our settings. Tab.~\ref{table:replicafull} demonstrates that our method significantly outperforms others in all metrics. Visual comparisons with NeuralRGBD in Fig.~\ref{fig:scannet-replica} (b) further highlight our more accurate reconstructions.

\begin{figure}[tb]
  \centering
   \includegraphics[width=0.9\linewidth]{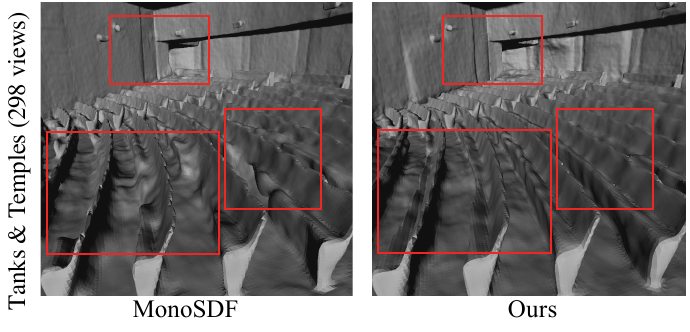}
\caption{\label{fig:tnt}Visual comparison with MonoSDF on Tanks and Temples.}
\vspace{-0.15in}
\end{figure}

\begin{figure}[tb]
  \centering
   \includegraphics[width=1.\linewidth]{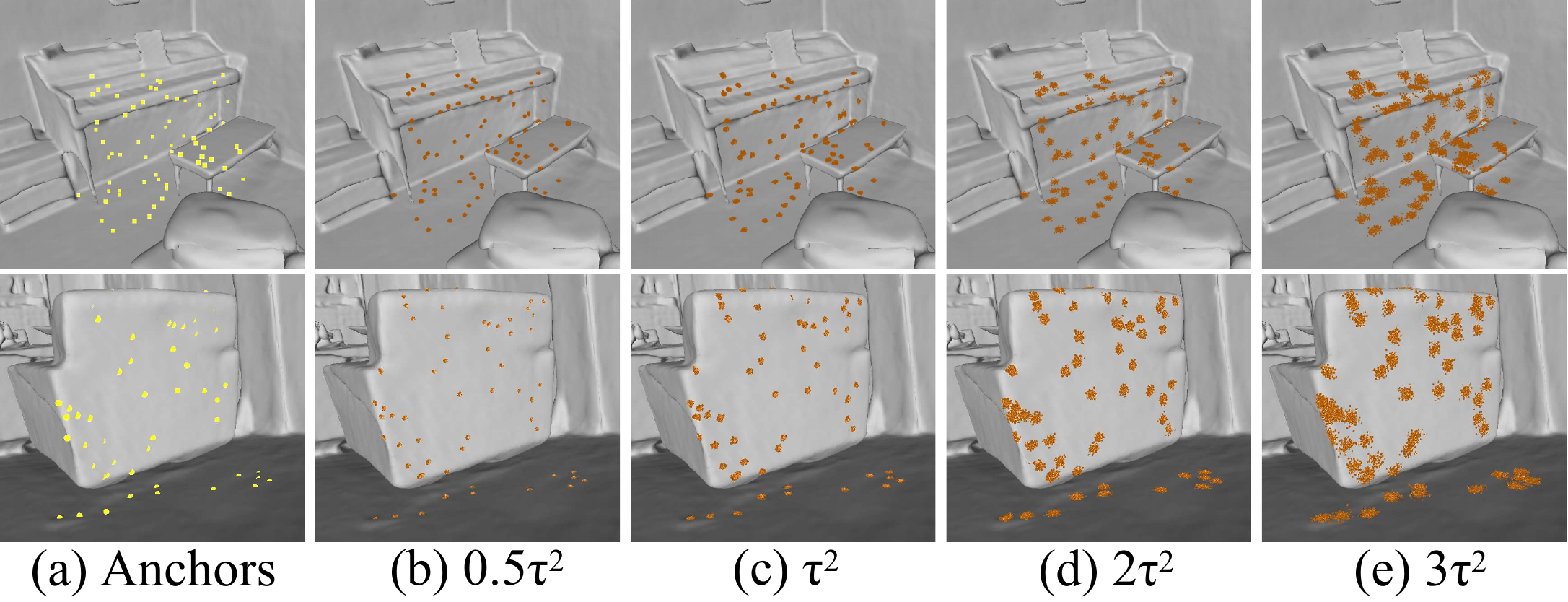}
\caption{\label{fig:Size}Comparisons on sizes of patch surfaces.}
   \vspace{-0.1in}
\end{figure}

\begin{table*}[tb]
\centering
\resizebox{0.8\linewidth}{!}{
\begin{tabular}{c c c c c c c c c c c}
\toprule
& &room\_0 & room\_1 & room\_2 & office\_0 & office\_1 & office\_2 & office\_3 & office\_4 & Mean\\
\midrule
\rowcolor{colorSep}{\textit{\textbf{RGB-D input}}}\\
\multirow{4}{*}{{NICE-SLAM}}
& Acc[cm]$\downarrow$ &  3.53 &  3.60 & 3.03 & 5.56 & 3.35 & 4.71 &  3.84 &  3.35 & 3.87 \\
& Comp[cm]$\downarrow$ & 3.40 & 3.62 &  3.27 &  4.55 & 4.03 &  3.94 &  3.99 & 4.15 &  3.87 \\
&Recall $\uparrow$ &  86.05 & 80.75 &  87.23 &  79.34 &  82.13 &  80.35 & 80.55 & 82.88 &  82.41 \\
&Normal C.$\uparrow$ &  91.92 &  91.36 & 90.79 & 89.30 &  88.79 &  88.97 & 87.18 & 91.17 & 89.93\\
\hline
\hline
\rowcolor{colorSep}{\textit{\textbf{RGB monocular input}}}\\
{\multirow{4}{*}{DROID-SLAM}}
&Acc[cm]$\downarrow$ &12.18& 8.35& 3.26& 3.01&2.39& 5.66& 4.49& 4.65& 5.50\\
&Comp[cm]$\downarrow$ &8.96& 6.07 &16.01 &16.19 &16.20& 15.56& 9.73& 9.63 &12.29\\
&Recall$\uparrow$ &60.07& 76.20 &61.62 &64.19& 60.63 &56.78& 61.95& 67.51& 63.62\\
&Normal C.$\uparrow$ & 72.81 &74.71 &79.21 &77.53& 78.57 &75.79& 77.69 &76.38 &76.59\\
\hline
{\multirow{4}{*}{NICER-SLAM}}
&Acc[cm]$\downarrow$ &\textbf{2.53}& 3.93& 3.40& 5.49 &3.45 &4.02 &3.34& 3.03 &3.65\\
&Comp[cm]$\downarrow$ &3.04 &4.10 &3.42& 6.09 &4.42& 4.29& 4.03 &3.87 &4.16
\\
&Recall $\uparrow$ & 88.75& 76.61& 86.10& 65.19& 77.84& 74.51& 82.01& 83.98& 79.37\\
&Normal C.$\uparrow$ & 93.00 &91.52& 92.38& 87.11& 86.79& 90.19& 90.10& 90.96 &90.27\\
\hline
{\multirow{4}{*}{OURS}}
&Acc[cm]$\downarrow$ &2.54 & \textbf{1.75}& \textbf{2.41}& \textbf{2.24}& \textbf{1.70}& \textbf{2.78} & \textbf{2.90} &\textbf{2.31} &\textbf{ 2.33}\\
&Comp[cm]$\downarrow$ &\textbf{2.08}& \textbf{2.54}& \textbf{2.58}& \textbf{2.89}& \textbf{3.11}& \textbf{2.94} & \textbf{3.11} & \textbf{2.77} & \textbf{2.75}\\
&Recall$\uparrow$ &\textbf{96.00}& \textbf{91.99} &\textbf{93.00}& \textbf{88.59}& \textbf{90.34} &\textbf{86.99}&\textbf{88.99}&\textbf{91.59}&\textbf{90.93}\\
&Normal C.$\uparrow$ & \textbf{95.60} &\textbf{94.20} &\textbf{94.36}& \textbf{90.47}& \textbf{92.79}& \textbf{92.40}&\textbf{92.19}&\textbf{94.49}&\textbf{93.31}\\
\hline
  \bottomrule
\end{tabular}}
\caption{Numerical comparison in each scene on Replica.}
\label{table:replicafull}
\end{table*}

\begin{figure*}[tb]
  \centering
   \includegraphics[width=\linewidth]{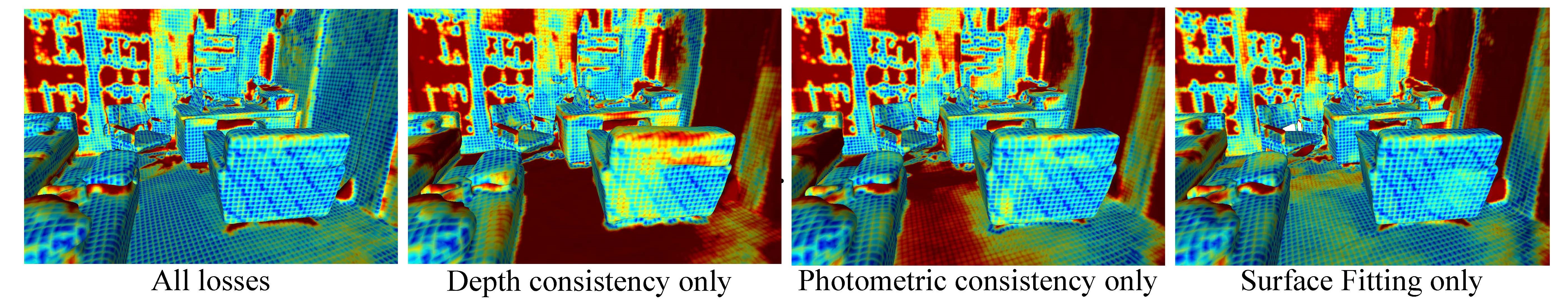}
\caption{\label{fig:abloss}Effect of losses. We use MonoSDF as a baseline and apply one loss each time.}
\end{figure*}

\noindent\textbf{Tanks and Temples. }We follow the same experimental setting in MonoSDF~\cite{Yu2022MonoSDF}, and apply monocular cues during optimization. Since the GT meshes are not publicly available, we only compare visual results with MonoSDF. The visual comparison in Fig.~\ref{fig:tnt} show that our method can recover more accurate geometry details. 

\noindent\textbf{Analysis. }For methods learning implicit representations from multi-view images, the most challenging problem is to infer the accurate zero level set of the implicit function. Our significant improvements over the latest methods mainly come from the fact that our surface constraints address this challenge quite well. We visualize across section of the reconstructed scenes and GT meshes in Fig.~\ref{fig:ground}. We can see that our surface (in yellow) is much closer to the GT mesh (in grey) than MonoSDF, although both MonoSDF and ours reconstruct complete and smooth surfaces. Since MonoSDF can not recover the zero level set accurately, its reconstructed surface is inflated, not only near the cross section but also other places in the scene, which leads to the floating surface over the GT mesh with large errors.

\begin{table}[tb]
\centering
\resizebox{\linewidth}{!}{
\begin{tabular}{c c c c c c c}
\toprule
$J$,$\tau^2$& Acc$\downarrow$ & Comp$\downarrow$ & CD-L1$\downarrow$ & Prec$\uparrow$ & Recall$\uparrow$ & F-score$\uparrow$\\
\midrule
1 & 0.037 &  0.040 &  0.039 & 0.781 & 0.738 & 0.760\\
\hline
9 & \textbf{0.036} & \textbf{0.039} & \textbf{0.037} & \textbf{0.820} & 0.777 & \textbf{0.797}\\
\hline
16 & 0.037&  0.039 &  0.038 & 0.802 & 0.754 & 0.777\\
\hline
25 & 0.037 &  0.041  &  0.039 & 0.789 & 0.732 & 0.759\\
\hline
Pixels & 0.037& \textbf{0.039}& 0.038 & 0.801 & \textbf{0.787} & 0.794\\
\midrule
$0.5\tau^2$ & 0.037 &  0.040 &  0.039 & 0.783 & 0.743 & 0.763\\
\hline
$\tau^2$ & \textbf{0.036} & \textbf{0.039} & \textbf{0.037} & \textbf{0.820} & \textbf{0.777} & \textbf{0.797}\\
\hline
$2\tau^2$ & \textbf{0.036} &  \textbf{0.039} &  \textbf{0.037} & 0.805 & 0.757 & 0.781\\
\hline
$3\tau^2$ & 0.037 &  0.040 &  0.038 & 0.793 & 0.738 & 0.766\\
\bottomrule
\end{tabular}}
\caption{Effect of point number $J$ and variance $\tau^2$.}
\label{table:pointnumber}
\end{table}

\subsection{Ablation Studies}
Ablation studies on ScanNet released by MonoSDF justify the effectiveness of modules in our method.

\noindent\textbf{Loss. }We first justify the effectiveness of losses in Tab.~\ref{table:losses}. We mainly focus on the losses for surface constraints, since the effectiveness of losses for volume rendering have been widely justified in previous studies. Compared to the baseline, each loss for a surface constraint can improve the performance. Fig.~\ref{fig:abloss} shows that each loss may improve different aspects, for example, the depth consistency makes the surface smoother, the photometric consistency makes the surface more compact, and the surface fitting loss also contributes to the smoothness of the surface.

\noindent We also try larger weights or smaller weights on all the three losses for surface constraints. The results of ``10$\times$'' and ``0.1$\times$'' show that weighting more or less on surface constraints do not balance well with the constraints through volume rendering and degenerate the reconstruction accuracy.

\noindent\textbf{Point Number $J$ in a Surface Patch $s$. }We explore the impact of point number $J$ on the performance. We try different densities  $J=\{1,9,16,25\}$ on surface patches by sampling queries using the same Gaussian distribution. The numerical comparison in Tab.~\ref{table:pointnumber} shows that a too small number of queries may not be able to represent a surface, such as $J=1$, while a too large number of points may not improve the performance further but increase the time complexity. We also compare our patch-level photo consistency with the pixel-level photo consistency. The results of ''Pixels'' show that patch-level photo consistency achieves better performance.

\noindent\textbf{Guassian Distribution for Sampling. }We also conduct an experiment to explore the impact of the variance $\tau^2$ of the Guassian distribution on the performance.  $\tau^2$ is initially set to be the distance between two other pixels in the 3D world coordinate, and use it as a baseline. We try different variance candidates including $\{0.5\tau^2,\tau^2,2\tau^2,3\tau^2\}$ to sample $J=9$ points around each anchor. The comparison in Tab.~\ref{table:pointnumber} shows that a small variance may not cover a large enough area which degenerates the performance of inference while a large variance covers a too large area where points may be ruled out by our masks. This may decrease the efficiency and also degenerate the performance. We visualize sizes of these patch surfaces in Fig.~\ref{fig:Size}, where all patch surfaces tightly locate on the reconstructed surfaces.

\begin{table}
\centering
\resizebox{\linewidth}{!}{
\begin{tabular}{c |c c c c c c}
\toprule
& Acc$\downarrow$ & Comp$\downarrow$ & CD-L1$\downarrow$ & Prec$\uparrow$ & Recall$\uparrow$ & F-score$\uparrow$\\
\midrule
Baseline & \textbf{0.035 }&  0.048 &  0.042 & 0.799 & 0.681 & 0.733\\
\hline
Only $L_{DC}$ & 0.039 &  0.041 &  0.040 & 0.778 & 0.753 & 0.766\\
\hline
Only $L_{NCC}$ & 0.036 &  \textbf{0.039} &  \textbf{0.037} & 0.801 & 0.744 & 0.772\\
\hline
Only $L_{Fit}$  & 0.037 &  \textbf{0.039} &  0.038 & 0.794 & \textbf{0.777} & 0.785\\
\hline
\hline
10$\times$ & 0.041 &  0.043 &  0.042 & 0.764 & 0.753 & 0.759\\
\hline
0.1$\times$ & 0.039 &  0.045 &  0.042 & 0.775 & 0.712 & 0.743\\
\hline
\hline
No $w_j$ & 0.038 &  \textbf{0.039} &  0.038 & 0.790 & 0.760 & 0.770\\
\hline
No $\eta_j$ & 0.036 &  0.040 &  0.038 & 0.801 & 0.768 & 0.784\\
\hline

No pulling & 0.037 &  0.041 &  0.039 & 0.772 & 0.734 & 0.753\\
\hline
Ours-Full & 0.036 & \textbf{0.039} & \textbf{0.037} & \textbf{0.820} & \textbf{0.777} & \textbf{0.797}\\
  \bottomrule
\end{tabular}}
\caption{Effect of Losses and pulling.}
\label{table:losses}
\end{table}

\noindent\textbf{Mask $w_j$ and Weight $\eta_j$. }We highlight the effect of the mask $w_j$ and the weight $\eta_j$ by removing one of them each time. The comparison is reported in Tab.~\ref{table:losses}. The decreased results ``No $w_j$'' and ``No $\eta_j$'' show that $w_j$ can rule out outliers during pulling while $\eta_j$ can make network focus more on the most important $p_j'$ on the patch surface.

\noindent\textbf{The effect of pulling. }We conduct an experiment to demonstrate that the pulling mechanism can boost the performance. We conduct this experiment without pulling by imposing depth, photometric consistency and surface fitting directly on 3D anchor $q$ rather than the sensed surface patch. As shown in Tab.~\ref{table:losses}, without pulling, the overall performance decreases. 


\section{Conclusion}
We propose a method to infer SDFs from multi-view images via volume rendering with surface patch sensing. By using SDF to define a local patch around the estimated ray-surface intersection, we are enabled to directly apply surface constraints in volume rendering. Leveraging gradients and signed distances, we pull sampled points onto the zero level set, enabling explicit surface constraints that enhance accuracy and capture finer geometry details. Numerical and visual comparisons demonstrate our superiority over the latest methods, validated across widely used benchmarks.

{
    \small
    \bibliography{papers}
}


\end{document}